\documentclass{tlp}
\usepackage{aopmath}
\usepackage{algorithm}
\usepackage{amssymb}
\usepackage{algpseudocode}
\usepackage{tikz}
\usepackage{multicol}
\usepackage{subfig}
\usepackage{listings}
\usepackage{url}
\usetikzlibrary{arrows,shapes,trees}
\newtheorem{definition}{Definition} % [section]
\newtheorem{example}{Example} % [section]
\def\tab{\mbox{\hspace{5mm}}}
\def\codeif{\texttt{:-}}
\def\ruleend{\texttt{.}}
\def\naf{\texttt{not }}
\newcommand{\asp}[1]{\mbox{\small $\mathtt{#1}$}}

\begin{document}

\title{Learning Weak Constraints in Answer Set Programming}

\author[M. Law, A. Russo, K. Broda]{ Mark Law, Alessandra Russo\thanks{This research is partially
funded by the 7th Framework EU-FET project 600792 ``ALLOW Ensembles'', and the
EPSRC project EP/K033522/1 ``Privacy Dynamics''.} 
, Krysia Broda
\\ Department of Computing, Imperial College London, SW7 2AZ\\
\email{$\lbrace$mark.law09, a.russo, k.broda$\rbrace$@imperial.ac.uk }}

\submitted {27 April 2015}
 \revised {3 July 2015}
 \accepted{21 July 2015}

\maketitle

\begin{abstract}
This paper contributes to the area of
%non monotonic
inductive logic programming
by presenting a new learning framework that allows the learning of weak
constraints in Answer Set Programming (ASP). The framework, called
\emph{Learning from Ordered Answer Sets}, generalises our previous work on
learning ASP programs without weak constraints, by considering a new notion of
examples as \emph{ordered} pairs of partial answer sets that exemplify which
answer sets of a learned hypothesis (together with a given background
knowledge) are \emph{preferred} to others. In this new learning task inductive
solutions are searched within a hypothesis space of normal rules, choice
rules, and hard and weak constraints.  We propose a new algorithm, % called
ILASP2, which is sound and complete with respect to our new learning framework.
We investigate its applicability to learning preferences in
% the setting of
an interview scheduling problem and also demonstrate that when restricted to
the task of learning ASP programs without weak constraints, ILASP2 can be much
more efficient than our previously proposed system.

\end{abstract}
\begin{keywords}
Non-monotonic Inductive Logic Programming,
Preference Learning,
Answer Set Programming
\end{keywords}

\vspace{-3.5mm}
\label{sec:introduction}
\section{Introduction}
\vspace{-1.5mm}

Preference Learning has received much attention over the last decade from
within the machine learning community. A popular approach to preference
learning is \emph{learning to rank}~\cite{furnkranz2003pairwise,geisler2001modeling}, 
where the goal is to learn to rank any two objects given some examples of pairwise
preferences (indicating that one object is preferred to another). Many of these
approaches use traditional machine learning tools such as neural networks~\cite{geisler2001modeling}.

On the other hand, the field of Inductive Logic Programming (ILP)
\cite{Muggleton1991} has seen significant advances in recent years, not only
with the development of systems, such as
\cite{Ray2004,Kimber2009,Corapi2010,Muggleton2011,muggleton2013meta}, but also
the proposals of new frameworks for learning
\cite{Otero2001,Sakama2009,JELIAILASP}.  In most approaches to ILP, a learning
task consists of a background knowledge $B$ and sets of positive and negative
examples. The task is then to find a hypothesis that, together with B, covers
all the positive examples but none of the negative examples.
While in previous work ILP systems such as TILDE~\cite{tilde} and
Aleph~\cite{srinivasan2001aleph} have been applied to preference
learning~\cite{dastani2001modeling,horvath2012model}, this has addressed
learning ratings, such as $good$, $poor$ and $bad$, rather than rankings over
the examples.  Ratings are not expressive enough if we want to find an optimal
solution as we may rate many objects as $good$ when some are better than
others.  Answer Set Programming (ASP), on the other hand, allows the expression
of preferences through \emph{weak constraints}. In a usual application of ASP,
one would write a logic program which has many answer sets, each corresponding
to a solution of the problem.  The program can also contain weak constraints
(or optimisation statements) which impose an ordering on the answer sets.
Modern ASP solvers, such as clingo~\cite{gekakaosscsc11a}, can then find the
optimal answer sets, which correspond to the optimal solutions of the problem.
For instance, in a scheduling problem, we could define an ASP program, whose
answer sets correspond to timetables, and weak constraints that represent
preferences over these timetables (see~\cite{timetable} for an example
application of ASP in timetabling).

In this paper, we propose a new learning framework, \emph{Learning from
Ordered Answer Sets} ($ILP_{LOAS}$), that allows the learning of ASP programs
with weak constraints. This framework extends the notion of learning from
answer sets proposed in~\cite{JELIAILASP}, where ASP programs without weak
constraints were learned using only positive and negative examples of partial
answer sets.  In our new learning task $ILP_{LOAS}$, additional examples are
defined, as ordered pairs of partial answer sets, and the language bias
captures a hypothesis space of ASP programs containing normal rules, choice
rules and hard and weak constraints. A new algorithm is presented and proved to
be sound and complete with respect to $ILP_{LOAS}$.

To demonstrate the applicability of our framework, we consider, as a
running example, an interview timetabling problem and the task of learning, as
weak constraints, academics' preferences for scheduling undergraduate
interviews.  An academic might be more comfortable interviewing for one course
than another, might prefer not to have many interviews on the same day, or
might hold  both of these preferences but regard the former as more important.
Given ordered pairs of partial timetables, our approach is able to learn these
preferences as weak constraints.
%, which guarantee that the timetables will be
%ordered to reflect these preferences.

The paper is structured as follows. Our new learning framework, $ILP_{LOAS}$ is
presented in Section~\ref{sec:loas}. It extends the notion of Learning from
Answer Sets~\cite{JELIAILASP} to the new task of learning weak constraints. We
discuss formal properties of the framework such as the complexity of deciding
the existence of a solution. Our learning algorithm \emph{ILASP2} is
described in Section~\ref{sec:alg}, together with experimental results based on
a scheduling example (Section~\ref{sec:experiments}). We also show that ILASP2
can have increased efficiency over our previous system when learning programs
without weak constraints. Discussion on related and future work concludes the
paper.

\vspace{-2.1mm}
\section{Background}
\vspace{-1.4mm}
\label{sec:background}

In this section we introduce the concepts needed in the paper.  Given any atoms
$h, h_1,\ldots,h_o, b_1,\ldots,b_n,c_1,\ldots,c_m$, $h \leftarrow b_1,\ldots,
b_n, \naf c_{1},\ldots, \naf c_{m}$ is called a \emph{normal rule}, with $h$ as
the \emph{head} and $b_1,\ldots, b_n, \naf c_{1},\ldots,\naf c_{m}$
(collectively) as the \emph{body} (``$\mathtt{not}$" represents negation as
failure); a rule $\leftarrow b_1,\ldots, b_n,$ $\naf c_{1},\ldots,$\break $\naf c_{m}$
is a \emph{hard constraint}; a \emph{choice rule} is a rule $l\{h_{1}, \ldots,
h_{o}\}u\leftarrow b_1,\ldots, b_n,$\break $\naf c_{1},\ldots, \naf c_{m}$ (where $l$
and $u$ are integers) and its head is called an \emph{aggregate}. A variable in
a rule $R$ is \emph{safe} if it occurs in at least one positive literal in the
body of $R$.  A program $P$ is assumed to be a finite set of normal rules,
choice rules, and hard constraints.  The Herbrand Base of $P$, denoted $HB_P$,
is the set of variable free (ground) atoms that can be formed from predicates
and constants in $P$. The subsets of $HB_P$ are called the (Herbrand)
interpretations of $P$.  A ground aggregate $l\{h_{1}, \ldots, h_{o}\}u$ is
satisfied by an interpretation $I$ iff $l\leq| I\cap\{h_{1},
\ldots,h_{o}\}|\leq u$.

As we restrict our programs to sets of normal rules, (hard) constraints and
choice rules, we can use the simplified definitions of the \emph{reduct} for
choice rules presented in \cite{simpleReduct}. Given a program $P$ and an
Herbrand interpretation $I \subseteq HB_{P}$, the reduct $P^{I}$ is constructed
from the grounding of $P$ in 4 steps: firstly, remove rules whose bodies
contain the negation of an atom in $I$; secondly, remove all negative literals
from the remaining rules; thirdly, replace the head of any hard constraint, or
any choice rule whose head is not satisfied by $I$ with $\bot$ (where
$\bot\notin HB_P$); and finally, replace any remaining choice rule $\lbrace
h_1,\ldots,h_m\rbrace\leftarrow b_1,\ldots,b_n$ with the set of rules $\lbrace
h_i \leftarrow b_1,\ldots,b_n \mid h_i \in I \cap \lbrace h_1,\ldots,
h_m\rbrace\rbrace$. Any $I \subseteq HB_{P}$ is an \emph{answer set} of $P$ if
it is the minimal model of the \emph{reduct} $P^{I}$.  Throughout the paper we
denote the set of answer sets of a program $P$ with $AS(P)$.

Unlike hard constraints in ASP, \emph{weak constraints} do not affect what is,
or is not, an answer set of a program $P$. Hence the above definitions also
apply to programs with weak constraints. Weak constraints create an ordering
over $AS(P)$ specifying which answer sets are ``better'' than others.  The set
of \emph{optimal} (best) answer sets of $P$ is denoted as $AS^*(P)$.  A
\emph{weak constraint} is of the form $:\sim b_1,\ldots,b_n, \naf c_1,\ldots,$
$\naf c_m\ruleend[w@l,t_1,\ldots, t_o]$ where $b_1, \ldots, b_n$,
$c_1,\ldots,c_m$ are atoms, $w$ and $l$ are terms specifying the \emph{weight}
and the \emph{level}, and $t_1,\ldots,t_o$ are terms. A weak constraint $W$ is
\emph{safe} if every variable in $W$ occurs in at least one positive literal in
the body of $W$.  At each \emph{priority level} $l$, the aim is to discard any
answer set which does not minimise the sum of the weights of the ground weak
constraints (with level $l$) whose bodies are true. The higher levels are
minimised first. Terms specify which ground weak constraints should be
considered unique.  For any program $P$ and $A\in AS(P)$, $weak(P,A)$ is the
set of tuples $(w,l,t_1,\ldots,t_o)$ for which there is some $:\sim
b_1,\ldots,b_n,\naf c_1,\ldots, \naf c_m\ruleend[w@l,t_1,\ldots,t_o]$ in the
grounding of $P$ such that $A$ satisfies $b_1,\ldots,b_n,\naf c_1,\ldots, \naf
c_m$.

We now give the semantics for weak constraints~\cite{ASPCORE2}.
%
%\begin{definition}
  For each level $l$, $P^{l}_A = \sum_{(w,l,t_1,\ldots,t_o)\in weak(P, A)} w$.
  For $A_1, A_2 \in AS(P)$, $A_1$ \emph{dominates} $A_2$ (written $A_1
  \succ_{P} A_2$) iff $\exists l$ such that $P^{l}_{A_1} < P^{l}_{A_2}$ and
  $\forall m > l, P^m_{A_1} = P^m_{A_2}$.  An answer set $A\!\in\! AS(P)$ is
  \emph{optimal} if it is not dominated by any $A_2\!\in\!AS(P)$.
%\end{definition}

\begin{example}\label{eg:timetable}

  Let $P$ be the program consisting of $\asp{slot(m,1)}$; $\asp{slot(m,2)}$;
$\asp{slot(t,1)}$;  $\asp{slot(t,2)}$; and $\asp{0\lbrace assign(D, S)\rbrace 1
\leftarrow slot(D,S)}$, which assigns 0 to 4 slots in a schedule
($\asp{slot(m,1)}$ represents slot 1 on Monday). Let $W_1$, $W_2$ and $W_3$ be
the weak constraints \break $\asp{:\sim assign(D, S)\ruleend[1@1]}$, $\asp{:\sim
assign(D, S)\ruleend[1@1, D]}$ and $\asp{:\sim assign(D, S)\ruleend[1@1, D,
S]}$ respectively.  Applying each weak constraint to $P$ gives as its optimal
answer set the one in which no slots are assigned. The remaining answer sets
are ordered in the following way: $W_1$ considers all schedules in which slots
have been assigned to be equally optimal, as there is only one unique set of
terms $t_1,\ldots,t_n$ which is the empty set; $W_2$ minimises the number of
days in which slots have been assigned, as there is one unique set of terms per
day; and finally, $W_3$ minimises the number of assignments made, as each
combination of day and slot has a unique set of terms.

\end{example}

\noindent
In an ILP task, the hypothesis space is often characterised by mode
declarations~\cite{Muggleton2011}.  A \emph{mode bias} can be defined as a pair
of sets of mode declarations $\langle M_{h}, M_{b}\rangle$, where $M_{h}$
(resp.  $M_{b}$) are the \emph{head} (resp. \emph{body}) declarations.  Each
mode declaration $m \in M_h$, or $m \in M_b$, is a literal whose abstracted
arguments are either $v$ or $c$.  An atom $a$ is \emph{compatible} with a mode
declaration $m$ if replacing the instances of $v$ in $m$ by variables, and the
instances of $c$ by constants yields $a$.  The search space is defined to be
the set of rules of the form $h\leftarrow b_1,\ldots, b_n, \naf c_{1},\ldots,
\naf c_{m}$ where (i) $h$ is empty, $h$ is an atom compatible with some $m\in
M_h$, or $h$ is an aggregate $l\{ h_1,\ldots, h_k \} u$ such that $0 \leq l
\leq u\leq k$ and $\forall i \in [1, k]$ $h_i$ is compatible with some $m \in
M_h$; (ii) $\forall i \in [1, n]$, $\forall j \in [1, m]$ $b_i$ and $c_{j}$ are
each compatible with at least one mode declaration in $M_{b}$; and finally
(iii) all variables in the rule are safe. We require the rules to be safe
because ASP solvers such as clingo~\cite{gekakaosscsc11a} have this
requirement.  We denote the search space defined by a given mode bias ($M_{h},
M_{b}$) as $S_{LAS}(M_{h}, M_{b})$.

In \cite{JELIAILASP}, we presented a new learning task, \emph{Learning from
Answer Sets} ($ILP_{LAS}$) which used \emph{partial interpretations} as
examples.  A partial interpretation $e$ is a pair $\langle e^{inc}, e^{exc}
\rangle$ of sets of ground atoms, called \emph{inclusions} and
\emph{exclusions}.  An answer set $A$ is said to \emph{extend} $e$ if and only
if $(e^{inc} \subseteq A) \land (e^{exc} \cap A = \emptyset)$. Given partial
interpretations $e_1$ and $e_2$,
 $e_1$ extends $e_2$ iff $e^{inc}_2\subseteq
e^{inc}_1$ and $e^{exc}_2\subseteq e^{exc}_1$.

\begin{definition} \label{def:LASTheory}
A \emph{Learning from Answer Sets} task is a tuple $T\!=\!\langle B,
S_{LAS}(M_h, M_b), E^{+}, E^{-} \rangle$ where $B$ is the background knowledge,
$S_{LAS}(M_h, M_b)$ is the search space defined by a bias $\langle M_h,
M_b\rangle$, $E^{+}$ and $E^{-}$ are sets of partial interpretations called the
positive and negative examples. A hypothesis $H$ is in $ILP_{LAS}(T)$, the set
of all \emph{inductive solutions} of $T$, if and only if $H\!\subseteq\!
S_{LAS}(M_h, M_b)$; $\forall e^{+}\!\in\!E^{+}$  $\exists A\!\in\!  AS(B\cup
H)$ such that $A$ extends $e^{+}$; and finally, $\forall e^{-}\!\in\!E^{-}$
$\nexists A\!\in\!AS(B\cup H)$ such that $A$ extends $e^{-}$.
\end{definition}

The task of ILP is usually to find  \emph{optimal} hypotheses, where optimality
is often defined by the number of literals in a hypothesis.  In $ILP_{LAS}$,
aggregates are converted to disjunctions before the literals are counted,
giving a higher ``cost'' for learning an aggregate; for example, $1 \lbrace p,
q \rbrace 2$ is converted to $(p\land \naf q)\lor (q \land \naf p)\lor (p\land
q)$ giving a length of 6. $ILP_{LAS}$ aims at learning ASP programs consisting
of normal rules, choice rules and hard constraints.  This paper extends this
notion to a new learning task capable of learning weak constraints.

\vspace{-1.2mm}
\section{Learning from Ordered Answer Sets} \label{sec:loas}
\vspace{-0.7mm}

%To learn weak constraints we extend the notion of mode
%bias with two new sets of mode declarations: $M_{o}$ and $M_w$. The
%former is used to specify what is allowed to appear in the body of a weak
%constraint, whereas the latter is used to specify what is allowed to appear as
%a weight. A positive integer $l_{max}$ is also given to indicate the maximum
%number of levels that can occur in $H$.
To learn weak constraints we extend the notion of mode bias with two new sets
of mode declarations: $M_{o}$ specifies what is allowed to appear in
the body of a weak constraint; whereas $M_w$ specifies what is allowed
to appear as a weight. A positive integer $l_{max}$ is also given to indicate
the number of levels that can occur in $H$.

\begin{definition} A mode bias with ordering is a tuple $M = \langle M_h, M_b,
M_{o}, M_{w}, l_{max} \rangle$, where $M_{h}$ and $M_{b}$ are respectively head
and body declarations, $M_{o}$ is a set of mode declarations for body literals
in weak constraints, $M_{w}$ is a set of integers and $l_{max}$ is a positive
integer. The search space $S_{M}$ is the set of rules $R$ that satisfy one of
the conditions:

  \begin{itemize}
    \item
      $R \in S_{LAS}(M_h, M_b)$.
    \item
      $R$ is a safe weak constraint $:\sim  b_1,\ldots, b_i,\naf
      b_{i+1},\ldots,\naf b_j\ruleend\lbrack w@l, t_1, \ldots, t_n \rbrack$
      such that $\forall k\!\in\![1, j]$ $b_k$ is compatible with $M_o$;
      $t_1,\ldots, t_n$ is the set of terms in $b_1, \ldots, b_j$; $w \in M_w$,
      $l \in [0, l_{max})$.
  \end{itemize}
\end{definition}

Note that even if we were to extend the learning task in
Definition~\ref{def:LASTheory} with this new notion of mode bias, such a task
would never have as its optimal solution a hypothesis which contains a weak
constraint. This is because a Learning from Answer Sets task has only examples
of what is, or is not, an answer set.  Any solution containing a weak
constraint $W$ will have the same answer sets without $W$, and would be
more optimal.
%shorter and therefore learned instead. 
We now define the notion of \emph{ordering
examples}.

\begin{definition} \label{def:Ordering} An \emph{ordering example} is a tuple
$o\!=\!\langle\!e_1, e_2\!\rangle$ where $e_1$ and $e_2$ are partial
interpretations.  An ASP program $P$ {\em bravely respects} $o$ iff $\exists
A_1, A_2\!\in\! AS(P)$ such that $A_1$ extends $e_1$, $A_2$ extends $e_2$ and
$A_1 \succ_{P} A_2$. $P$ {\em cautiously respects} $o$ iff $\forall A_1,
A_2\!\in\! AS(P)$ such that $A_1$ extends $e_1$ and $A_2$ extends $e_2$, it is
the case that $A_1\succ_{P} A_2$.
\end{definition}

\begin{example}\label{eg:mexample}
 Consider the partial interpretations $e_1=\langle\lbrace
 \asp{assign(m, 1), assign(m, 2)}\rbrace, \lbrace\asp{assign(t,1)},$
 $\asp{assign(t,2)}\rbrace\rangle$ and $e_2=\langle\lbrace\asp{assign(m,1),
 assign(t,1)}\rbrace, \emptyset\rangle$. Let $o=\langle e_1,
 e_2\rangle$ be an ordering example and recall $P$ and $W_1,\ldots,W_3$ from
 example~\ref{eg:timetable}. The only answer set of $P$ that extends $e_1$ is
 $m_1m_2$ (where $m_1m_2$ denotes $\lbrace \asp{assign(m, 1),
 assign(m,2)}\rbrace$), whereas the answer sets that extend $e_2$ are $m_1t_1,
 m_1m_2t_1, m_1m_2t_1t_2$ and $m_1t_1t_2$.  $P\cup W_1$ does not bravely or
 cautiously respect $o$ as it gives to all these answer sets the same
 optimality; $P\cup W_2$ both bravely and cautiously respects $o$, as each
 pair of answer sets extending the partial interpretations is ordered correctly
 (i.e.\ answer sets extending $e_{1}$ have slots allocated in only one day
 whereas all the answer sets extending $e_{2}$ have slots assigned in two
 days).  Finally, $P\cup W_3$ respects $o$ bravely but not cautiously
 (the pair of answer sets $m_1m_2$ and $m_1t_1$ is such that $m_1m_2
 \not\succ_{P} m_1t_1$).
 \end{example}

\noindent We can now define the notion of \emph{Learning from Ordered Answer
Sets} ($ILP_{LOAS}$).

\begin{definition} \label{def:LOASTheory} A \emph{Learning from Ordered Answer
Sets} task is a tuple $T=\langle B, S_M, E^{+}, E^{-}, O^{b}, O^{c} \rangle$
where $B$ is an ASP program, called the background knowledge, $S_{M}$ is the
search space defined by a mode bias with ordering $M$, $E^{+}$ and $E^{-}$ are
sets of partial interpretations called, respectively, positive and negative
examples, and $O^{b}$ and $O^{c}$ are sets of ordering examples over $E^{+}$
called brave and cautious orderings. A hypothesis $H \subseteq S_M$ is in
$ILP_{LOAS}(T)$, the \emph{inductive solutions} of $T$, if and only if:

  \begin{enumerate}
    \item
      Let $M_{h}$ and $M_{b}$ be as in $M$ and $H'$ be the subset of $H$ with
      no weak constraints.  $H' \in ILP_{LAS}(\langle B, S_{LAS}(M_h, M_b),
          E^{+}, E^{-}\rangle)$
    \item $\forall o \in O^b$ $B\cup H$ bravely respects $o$
    \item $\forall o \in O^c$ $B\cup H$ cautiously respects $o$
  \end{enumerate}
\end{definition}

\noindent
The notion of an optimal inductive solution of an $ILP_{LOAS}$ task is the same
as in $ILP_{LAS}$, with each weak constraint $W$ counted as the number of
literals in the body of $W$.  Note that the orderings are only over $E^{+}$
rather than orderings over any arbitrary partial interpretations. We chose to
make this restriction as we could not see a reason why a hypothesis would need
to respect orderings which are not extended by any pair of answer sets of
$B\cup H$. Note also that in the case where $O^b$ and $O^c$ are both empty, the
task reduces to an $ILP_{LAS}$ task.

\begin{example}\label{eg:scheduleex} 
Consider the $ILP_{LOAS}$ task $T$ with the following
background knowledge:

\begin{math}
  B=\left\{\begin{array}{l}
      \asp{slot(m, 1\ruleend\ruleend3)\ruleend slot(t, 1\ruleend\ruleend3)\ruleend
      slot(w, 1\ruleend\ruleend3)\ruleend}\\
      \asp{neq(1, 2)\ruleend neq(1, 3)\ruleend neq(2, 1)\ruleend neq(2, 3)\ruleend neq(3, 1)\ruleend  neq(3, 2)\ruleend}\\
      \asp{neq(m, t)\ruleend neq(m, w)\ruleend neq(t, m)\ruleend neq(t, w)\ruleend neq(w, m)\ruleend  neq(w, t)\ruleend}\\
      \asp{type(m,1,c1)\ruleend type(m,2,c2)\ruleend type(m,3,c2)\ruleend type(t,1,c2)\ruleend}\\
      \asp{type(t,2,c2)\ruleend type(t,3,c2)\ruleend type(w,1,c2)\ruleend type(w,2,c1)\ruleend type(w,3,c2)\ruleend}\\
      \asp{0 \lbrace assign(X, Y) \rbrace 1 \codeif slot(X, Y)\ruleend}\\
  \end{array}\right\}
\end{math}

Using the notation from example~\ref{eg:mexample}, let $T$ have the
positive examples $e_1=\langle \emptyset,m_2m_3t_1t_3w_1w_2\rangle$,
$e_2=\langle m_1m_2,\emptyset\rangle$, $e_3=\langle\emptyset,
m_1t_2w_1w_2\rangle$, $e_4=\langle t_1t_2t_3, \emptyset\rangle$, $e_5= \langle
m_2m_3t_1t_2t_3w_1w_3, \emptyset\rangle$; $e_6=\langle
m_1w_1w_3,\;m_2m_3t_1t_2t_3w_2\rangle$; two cautious orderings: $\langle e_1,
e_2\rangle$ and $\langle e_3, e_4\rangle$; and one brave ordering: $\langle
e_5, e_6\rangle$.  Consider $S_M$ to be defined by the mode declarations: $M_h
= M_b = \emptyset$; $M_o = \lbrace \asp{assign(v,v)},$ $\asp{neq(v,v),}$
$\asp{type(v,c)}\rbrace$; $M_w =\lbrace -1, 1\rbrace$; and $l_{max} =
2$.  Note that as each positive example is already covered by the background
knowledge and there are no negative examples, it remains to find a set of weak
constraints which meet conditions 2 and 3 of definition~\ref{def:LOASTheory}.
 One inductive solution $H$ of $T$ is $\lbrace \asp{:\sim
assign(D,S1),}$ $\asp{assign(D,S2),}
\asp{neq(S1,S2)\ruleend[1@1,D,S1,S2]}$\break $\asp{:\sim assign(D,S),}
\asp{type(D,S,c1)\ruleend[1@2,D,S]}\rbrace$; this respects the first cautious
ordering example because any timetable extending $e_1$ has at most one
$\asp{c1}$ course whereas $e_2$ has at least one, so $e_1$ is better or equal
to $e_2$ on the highest priority weak constraint; even if they are equal, a
timetable extending $e_1$ has at most one assignment per day and is, therefore,
always better on the lower priority weak constraint. $H$ also
respects the other cautious ordering and the timetables $m_2m_3t_1t_2t_3w_1w_3$
and $m_1w_1w_3$ correspond to answer sets which demostrate that the brave
ordering is respected.

In fact, there is no shorter hypothesis which meets conditions 1 to 3 and so
$H$ is an optimal inductive solution; moreover, the other optimal solutions are
equivalent hypotheses such as: $\lbrace\asp{:\sim
assign(D,S1),}\asp{assign(D,S2),}$ $\asp{neq(S1,S2)\ruleend[1@1,D,S1,S2]};\;\;$
$\asp{:\sim assign(D,S),}\asp{\naf
type(D,S,c2)\ruleend[1@2,D,S]}\rbrace$.  These hypotheses represent
the
% academic's
 preferences described in the introduction. They express that the
highest priority is to minimise the interviews for $c1$, and then to minimise
the slots in any one day.

%Using the notation from example~\ref{eg:mexample}, consider the task with
%positive examples $e_1=\langle \emptyset,\break $\lbrace\asp{assign(m,2)},$
%  $ \asp{assign(m,3)},$ $\asp{assign(t,1)},$  $\asp{assign(t,3)},$
%  $\asp{assign(w,1)},$\break $\asp{assign(w,2)}\rbrace\rangle$; $e_2=\langle
%  \lbrace\asp{assign(m,1)},$ $\asp{assign(m,2)}\rbrace,\emptyset\rangle$;
%  $e_3=\langle\emptyset, \lbrace\asp{assign(m,1)},$\break $\asp{assign(t,2),}$
%  $\asp{assign(w,1)},$ $\asp{assign(w,2)}\rbrace\rangle$; $e_4=\langle
%  \lbrace\asp{assign(t,1)},$ $\asp{ assign(t,2)},$
%  \break$\asp{assign(t,3)}\rbrace,$ $\emptyset\rangle$; $e_5=
%  \langle\lbrace\asp{assign(m,2)},$ $\asp{assign(m,3)},$
%  $\asp{assign(t,1)},$ $\asp{assign(t,2)},$\break $\asp{assign(t,3)},$
%  $\asp{assign(w,1)},$ $\asp{assign(w,3)}\rbrace,$ $\emptyset\rangle$;
%  $e_6=\langle \lbrace\asp{assign(m,1)},$ $\asp{assign(w,1)},$\break
%  $\asp{assign(w,3)}\rbrace, \lbrace\asp{assign(m,2)},$ $\asp{assign(m,3)},$
%  $\asp{assign(t,1)},$ $\asp{assign(t,2)}$, $\asp{assign(t,3)},$\break
%  $\asp{assign(w,2)}\rbrace\rangle\rbrace$; two cautious orderings: $\langle
%  e_1, e_2\rangle$ and $\langle e_3, e_4\rangle$; and one brave ordering:
%  $\langle e_5, e_6\rangle$.  Consider $S_M$ to be defined by the mode
%  declarations: $M_h = M_b = \emptyset$; $M_o = \lbrace \asp{assign(v,v)},$
%  $\asp{neq(v,v),}$ $\asp{type(v,c)}\rbrace$; $M_w =\lbrace -1, 1\rbrace$; and
%  finally, $l_{max} = 2$.  Then one optimal (shortest) inductive solution is: 

\end{example}

%\subsection{Theoretical Results}
%
%
%In this section, we discuss some of the formal properties of $ILP_{LOAS}$;
%specifically, we investigate the complexity of deciding whether an $ILP_{LOAS}$
%task has any solutions. All learning tasks in this section are assumed to be
%propositional ($B$ and $S_M$ are both ground). The proofs for the theorems can
%be found in~\cite{ILASP2Proof}.
We now discuss some of the formal properties of $ILP_{LOAS}$.  All learning
tasks in the rest of this section are assumed to be propositional ($B$ and
$S_M$ are both ground). The proofs for Theorems~\ref{thm:suff} to
\ref{thm:complexity} can be found in~\cite{ILASP2Proof}.
Theorems~\ref{thm:suff} and~\ref{thm:ness} state sufficient and necessary
conditions for there to exist solutions for an $ILP_{LOAS}$ task with an
unrestricted search space (hypotheses can be any set of normal rules, choice
rules and hard and weak constraints).

\begin{theorem}\label{thm:suff}
  Let $T$ be the $ILP_{LOAS}$ task $\langle B, E^{+}, E^{-}, O^{b},
  O^{c}\rangle$. The following conditions (in conjunction) are sufficient for
  there to exist solutions of $T$: (i) $\forall e \in E^{+}$, there
  is at least one model of $B$ which extends $e$; (ii) $\forall e_1\in E^{+}$,
  $\nexists e_2 \in (E^{+} \cup E^{-})$ such that $e_1$ extends $e_2$; (iii)
  there is no cyclic chain of ordering examples (in $O^b\cup O^c$) $\langle
  e_1, e_2\rangle, \langle e_2, e_3\rangle, \ldots,$ $\langle e_{n-1},
  e_n\rangle, \langle e_n, e_1\rangle$.
\end{theorem}

\begin{theorem}\label{thm:ness}
  Let $T$ be the $ILP_{LOAS}$ task $\langle B, E^{+}, E^{-}, O^{b},
  O^{c}\rangle$. The following conditions are necessary for there to exist
  solutions of $T$: (i) $\forall e \in E^{+}$, there is at least
  one model of $B$ which extends $e$; (ii) $\forall e_1\in E^{+}$, $\nexists
  e_2 \in E^{-}$ such that $e_1$ extends $e_2$; (iii) there is no cyclic chain of
  cautious orderings, $\langle e_1, e_2\rangle, \langle e_2,
  e_3\rangle, \ldots,$ $\langle e_{n-1}, e_n\rangle, \langle e_n, e_1\rangle$.
\end{theorem}

Note that if we consider the usual setting where hypotheses come from a search
space, the conditions in theorem~\ref{thm:ness} are still necessary, but the
conditions in theorem~\ref{thm:suff} are no longer sufficient as, even if the
conditions hold, the search space may be too restrictive.
Theorem~\ref{thm:complexity} states the complexity of deciding the existence of
solutions for both $ILP_{LAS}$ and $ILP_{LOAS}$ tasks. The interesting property
here is that deciding the existence of solutions for $ILP_{LOAS}$ is in the
same complexity class as $ILP_{LAS}$.

\begin{theorem}\label{thm:complexity}
  Let $T$ be any $ILP_{LAS}$ or $ILP_{LOAS}$ task.  Deciding whether $T$ has at
  least one inductive solution is $NP^{NP}$-complete.
\end{theorem}

\section{Algorithm}

\label{sec:alg}

We now describe our new algorithm, \emph{ILASP2}, capable of computing
inductive solutions of any $ILP_{LOAS}$ task, and present its soundness and
completeness results with respect to the notion of Learning from Ordered Answer
Sets task given in Definition~\ref{def:LOASTheory}. We omit the proofs of the
theorems in this paper, but they can be found in full in~\cite{ILASP2Proof}.
For details of how to download and use our prototype implementation of the
ILASP2 algorithm, see~\cite{ILASP_system}.

 ILASP2 extends
the concepts of \emph{positive} and \emph{violating} hypotheses, first
introduced in our previous algorithm ILASP \cite{JELIAILASP}, to cater for the
new notion of ordering examples. A hypothesis is said to
be \emph{positive} if it covers all positive examples and bravely respects all
the brave ordering examples. A positive hypothesis is defined to be
\emph{violating} if it covers at least one negative example or if it does not
respect at least one of the cautious ordering examples. These two notions are
formalised by Definitions~\ref{def:newposhyp} and~\ref{def:newviohyp}.

\begin{definition}\label{def:newposhyp}
  Let $T = \langle B, S_M$, $E^{+}$, $E^{-}, O^{b}, O^{c} \rangle$ be an
  $ILP_{LOAS}$ task. Any $H\subseteq S_M$ is a {\em positive hypothesis} iff
  $\forall e \in E^{+}$ $\exists A \in AS(B\cup H)$ such that $A$
  extends $e$, and $\forall o \in O^b$ $H\cup B$ bravely respects $o$.  The set
  of positive hypotheses of $T$ is denoted $\mathcal{P}(T)$.
\end{definition}

\begin{definition}\label{def:newviohyp}
  A positive hypothesis $H$ is a \emph{violating hypothesis} of $T = \langle B,
  S_M$, $E^{+}$, $E^{-}, O^{b}, O^{c} \rangle$, written $H\!\in\! \mathcal{V}(T)$,
  iff at least one of the following cases is true:
  \begin{itemize}
    \item
      $\exists e^{-}\!\in\!E^{-}$ and $\exists A\!\in\!AS(B\cup H)$ such that $A$ extends
      $e^{-}$. In this case we call $A$ a violating interpretation of $T$ and
      write $\langle H, A\rangle \in \mathcal{VI}(T)$.
    \item
      $\exists A_1, A_2\!\in\! AS(B\cup H)$ and $\exists\langle e_1, e_2\rangle
      \in O^c$ such that $A_1$ extends $e_1$, $A_2$ extends $e_2$ and
      $A_1\not\succ_{P} A_2$ with respect to $B\cup H$. In this case, we call
      $\langle A_1, A_2\rangle$ a \emph{violating pair} of $T$ and write
      $\langle H, \langle A_1, A_2\rangle\rangle \in \mathcal{VP}(T)$.
  \end{itemize}
\end{definition}

\vspace{-0.8mm}

\begin{example}\label{eg:manyvio}
  Consider an $ILP_{LOAS}$ task with $B$ equal to $P$ in
Example~\ref{eg:timetable} but with one additonal fact $\asp{busy(1,1)}$;
positive examples $e^{+}_{1}=$ $\langle\lbrace \asp{assign(t,1)},$
$\asp{assign(t,2)}\rbrace$, $\lbrace \asp{assign(m, 2)}\rbrace\rangle$ and
$e^{+}_{2}=$ $\langle\lbrace \asp{assign(m,2)}$, $\asp{assign(t,1)}\rbrace$,
$\emptyset\rangle$; one negative example $e^{-}_1=\langle\lbrace \asp{assign(m,
1)} \rbrace$, $\emptyset\rangle$; and one cautious ordering $\langle e^{+}_{1},
e^{+}_{2}\rangle$.  $S_M$ is unrestricted
%\footnote{
(hypotheses can be
constructed from any predicate that appears in $B$ and $E$). Three example
hypotheses are given below. Note that when we describe answer sets we omit the
facts in $B$.
%}.

  $H_1\!=\!\emptyset\!\in\!\mathcal{P}(T)$ as $e^{+}_{1}$ and $e^{+}_{2}$ are
covered and $O^{b}$ is empty; however, $H_{1}\in\mathcal{V}(T)$ for two
reasons: firstly it has a violating interpretation ($\lbrace
\asp{assign(m,1)}\rbrace$); secondly it has a violating pair ($\langle\lbrace
\asp{assign(t,1), assign(t,2)}\rbrace,\lbrace \asp{assign(m,2),
assign(t,1)}\rbrace\rangle$).

  $H_2=\lbrace\leftarrow\!\asp{busy(D, S),assign(D,S)}\rbrace$
  $\in\!\mathcal{P}(T)$.  $H_2$ has no violating interpretations, but it has a
  violating pair ($\langle \lbrace \asp{assign(t,1), assign(t,2)}\rbrace,$
  $\lbrace \asp{assign(m,2)},$ $\asp{assign(t,1)}\rbrace\rangle$).

    $H_3\!=\!H_2\!\cup\!\lbrace \asp{:\sim assign(D,
    S)\ruleend[1@1,D]}\rbrace\!\in\!\mathcal{P}(T)$.
    $H_3\!\notin\!\mathcal{V}(T)$, as it has no violating interpretations and
    its weak constraints minimise the days assigned (so it cautiously respects
    the ordering example).  It is, therefore, an inductive solution of the
    task.
\end{example}

One approach to computing the inductive solutions of an $ILP_{LOAS}$ task would
be to extend the original ILASP method with our new notions of positive and
violating hypotheses. Given a positive integer $n$, ILASP worked by
constructing an ASP meta representation of an $ILP_{LAS}$ task $T$, called the
\emph{task program} $T^n_{meta}$, whose answer sets could be mapped back to the
positive hypotheses of $T$ of length $n$. $T^n_{meta}$ could then be augmented
with an extra
constraint so that its answer sets corresponded exactly to the violating
hypotheses of length $n$. ILASP first computed the violating hypotheses of
length $n$, and then converted each of these to a constraint at the meta-level
(ruling out that hypothesis). When $T^n_{meta}$ was then augmented with these new
constraints, its answer sets corresponded exactly to the positive hypotheses
which were not computed the first time - the inductive solutions of length $n$.

The problem is that, in general, there can be many violating hypotheses which
are shorter than the first inductive hypothesis and ILASP will compute all of
them and add them into the task program as individual constraints. This
scalability issue would be worsened if we were considering adding weak
constraints to the search space. To overcome this, $ILASP2$ adopts a different
strategy: it eliminates \emph{classes} of hypothesis, i.e.\ hypotheses that are
violating for the same ``\emph{reason}'', namely they give rise to a particular
violating interpretation or a particular violating pair of interpretations.
The idea underlying the ILASP2 algorithm is to make use of two sets $VI$ and
$VP$ which accumulate, respectively, violating interpretations and violating
pairs of interpretations that are constructed during the search. We start
initially with two empty sets $VI$ and $VP$ and continually compute the set of
optimal \emph{remaining hypotheses} which do not violate any of the
\emph{reasons} in $VI$ or $VP$.  If a computed hypothesis gives rise to a new
violating interpretation then this interpretation is added to $VI$, if it gives
rise to a new violating pair of interpretations then this pair is added to
$VP$. If no optimal remaining hypotheses are violating, then these hypotheses
are the optimal inductive solutions of the task.

\begin{definition} \label{def:remainingViolatingReason}
  Let $T$ be an $ILP_{LOAS}$ task, $VI$ and $VP$ (resp.) be sets of violating
  interpretations and pairs of interpretations, and $B$ be the background
  knowledge. Any $H\in \mathcal{P}(T)$ is a \emph{remaining hypothesis} of $T$
  with respect to $VI\cup VP$ iff $VI \cap AS(B\cup H) = \emptyset$ and $\forall
  \langle I_1, I_2\rangle\in VP$ if $I_1, I_2 \in AS(B\cup H)$ then $I_1
  \succ_{B\cup H} I_2$. A remaining hypothesis $H$ is a \emph{remaining
  violating hypothesis} iff $\exists R$ such that $\langle H, R \rangle \in
  \mathcal{VI}(T)\cup \mathcal{VP}(T)$.
\end{definition}

\noindent We use an ASP meta-level representation to solve our search for
remaining hypotheses. As we rule out classes of hypothesis at the same time
(rather than using one constraint per violating hypothesis), our meta-level
representation is slightly more complex than that used in the original ILASP.
Due to this complexity, we define this representation in the online
appendix and give here the underlying intuition.

The intuition of our meta encoding is that for a given task $T$, we construct
an ASP program $T_{meta}$ whose answer sets can be mapped back to the positive
hypotheses of $T$. Given an answer set $A$ of $T_{meta}$ we write
$\mathcal{M}^{-1}_{hyp}(A)$ to denote the hypothesis represented by $A$. Each
positive hypothesis may be represented by many answer sets of $T_{meta}$ but if
this hypothesis gives rise to a violating interpretation, then at least one of
these answer sets will contain a special atom $v\_i$. If the hypothesis gives
rise to a violating pair of interpretations then at least one of the answer
sets of $T_{meta}$ representing the hypothesis will contain a special atom
$v\_p(t_1, t_2)$, where $\langle t_1, t_2\rangle$ is a pair of identifiers
corresponding to the cautious ordering example which is being violated.  There
is only one priority level in $T_{meta}$ and the optimality of its answer sets
is $2*|H|+1$ if the answer set does not contain the atom $violating$ ($violating$ is
defined to be true if and only if $v\_i$ or at least one $v\_p(t_1, t_2)$ is true) and
$2*|H|$ if it does.  This means that for any hypothesis $H$, the answer sets
corresponding to $H$ that do contain $violating$ are preferred to those which
do not.

We can use $T_{meta}$ to find optimal positive hypotheses of $T$. If these
positive solutions are violating, then the optimal answer sets will contain
$violating$. We can then rule these hypotheses out. We can extract violating
interpretations and violating pairs of interpretations from answer sets of
$T_{meta}$, using the functions $\mathcal{M}^{-1}_{vi}$ and
$\mathcal{M}^{-1}_{vp}$ respectively. Violating interpretations and violating
pairs of interpretations are both called \emph{violating reasons}. For any set
of violating reasons $VR=VI\cup VP$, we then have a second meta encoding
$VR_{meta}(T)$ which, when added to $T_{meta}$, rules out any hypotheses which
are violating for a reason already in $VR$. This means that the answer sets of
$T_{meta}\cup VR_{meta}(T)$ will represent the set of remaining hypotheses of
$T$ with respect to $VR$. These properties are guaranteed by
Theorem~\ref{thrm:vmeta}.

\begin{theorem}\label{thrm:vmeta}
  Given an $ILP_{LOAS}$ task and a set of violating reasons $VR$, let $AS$ be
  the set of optimal answer sets of $T_{meta}\cup VR_{meta}(T)$.
\begin{itemize}
\item
  If $\exists A \in AS$ such that $violating\!\in\!A$ then the set of optimal remaining
  violating hypotheses $VH$ is non empty and is equal to the set
  $\lbrace \mathcal{M}^{-1}_{hyp}(A) \mid A \in AS\rbrace$.
\item
  If no $A\in AS$ contains $violating$, then the set of optimal remaining
  hypotheses (none of which is violating) is equal to the set $\lbrace
  \mathcal{M}^{-1}_{hyp}(A) \mid A \in AS\rbrace$.
  \end{itemize}
\end{theorem}

\vspace{-1.5mm}
\begin{algorithm}
  \begin{algorithmic}
    \Procedure{ILASP2}{$T$}
    \State{$VR = []$}
    \State{$solution = solve(T_{meta}\cup VR_{meta}(T))$}
    \While{$solution \neq \mathtt{nil}$ \&\& $solution\ruleend optimality \% 2 == 0$}
      \State{$A = solution\ruleend answer\_set$}
      \If{$v\_i \in A$}
      \State{$VR\; +\!=\; \mathcal{M}^{-1}_{vi}(A)$}
      \ElsIf{$\exists t_1, t_2$ such that $v\_p(t_1, t_2) \in A$}
      \State{$VR\; +\!=\; \mathcal{M}^{-1}_{vp}(A)$}
      \EndIf
      \State{$solution = solve(T_{meta}\cup VR_{meta}(T))$}
    \EndWhile
    \State{\textbf{return}\,$\lbrace \mathcal{M}^{-1}_{hyp}(A)\mid A\!\in\! AS^{*}(T_{meta}\cup VR_{meta}(T))\rbrace$}
  \EndProcedure
\end{algorithmic}
\caption{ILASP2 \label{alg:ILASP2}}
\end{algorithm}
\vspace{-1.5mm}

\noindent Algorithm~\ref{alg:ILASP2} is the pseudo code of our algorithm
ILASP2.  It makes use of our meta encodings $T_{meta}$ and $VR_{meta}(T)$.  For
any program $P$, $solve(P)$ is a function which, in the case that $P$ is
satisfiable, returns a pair consisting of an optimal answer set together with
its optimality (as there is only one priority level in our meta encoding this
is treated as an integer); if $P$ is unsatisfiable then $solve(P)$ returns
$\mathtt{nil}$.  While there are optimal remaining violating hypotheses, ILASP2
finds them and records the appropriate violating reasons.  When there are no
optimal remaining hypotheses which are violating then either the
meta program will be unsatisfiable or the optimality of the optimal answer sets
will be odd (as the optimality of any $A\in AS(T_{meta} \cup VR_{meta}(T))$ is
$2*|\mathcal{M}^{-1}_{hyp}(A)|$ if $A$ contains $\asp{violating}$ and
$2*|\mathcal{M}^{-1}_{hyp}(A)|+1$ if not), and so ILASP2 stops and returns the
set of optimal remaining hypotheses. Theorem~\ref{thrm:sac} shows that ILASP2
is sound and complete with respect to the optimal inductive solutions of an
$ILP_{LOAS}$ task.  This result relies on the termination of $ILASP2(T)$, which
is guaranteed if $B\cup S_M$ grounds finitely.

\begin{theorem}\label{thrm:sac}
  Let $T$ be an $ILP_{LOAS}$ task. If $ILASP2(T)$ terminates, then
  $ILASP2(T)$ returns the set of optimal inductive solutions of
$ILP_{LOAS}(T)$.
\end{theorem}

\vspace{-4.0mm}
\section{Experiments}
\label{sec:experiments}

Although there are benchmarks for ASP solvers~\cite{ASPCOMP}, there are no
benchmarks for learning ASP programs. In~\cite{JELIAILASP} we discussed the
example of learning an ASP program with no weak constraints, representing the
rules of sudoku. Using the examples from the paper and a small search space
with only $283$ rules, the original ILASP algorithm takes $486\ruleend 2$s to
solve the task. This is due to the scalability issues discussed in
section~\ref{sec:alg} as there are $332437$ violating hypotheses found before
the first inductive solution. For the same task with ILASP2, there are only $9$
violating reasons found before the first inductive solution, meaning that
ILASP2 takes only $0\ruleend 69$s to solve the task.

As this is the first work on learning weak constraints, there are no existing
benchmarks suitable for testing our approach of learning from ordered answer
sets. We have,
therefore, further investigated the interview scheduling example discussed
throughout the paper.  Our experiments, in particular, test whether
$ILP_{LOAS}$ can successfully learn weak constraints from examples of brave and
cautious orderings. For the purpose of presentation, we assume our hypothesis
space, $S_M$, to be defined by the mode declarations:
$M_h\!=\!M_b\!=\!\emptyset$; $M_o=\asp{\lbrace assign(v,v), neq(v,v), type(v,c)
\rbrace}$; $M_w\!=\lbrace -1, 1\rbrace$; and finally, $l_{max} = 2$. We place
several restrictions on the search space in order to remove equivalent rules.
The size of $S_M$ is 184
%\footnote{
(our hypotheses can be any
subset of these 184 rules, so even considering only hypotheses with up to 3
rules this gives over a million different hypotheses).
%}.
The learning task uses background knowledge $B$ from Example~\ref{eg:scheduleex}.
As $S_M$ only contains weak constraints, for any $H\subseteq
S_M$, $AS(B \cup H)=AS(B)$. The learning tasks described in these experiments
therefore correspond to learning to rank the answer sets of $B$.

\begin{figure}[ht!]
  \centering
  \includegraphics[trim=0cm 0.1cm 0cm 0.1cm, clip=true, width=0.8\textwidth]{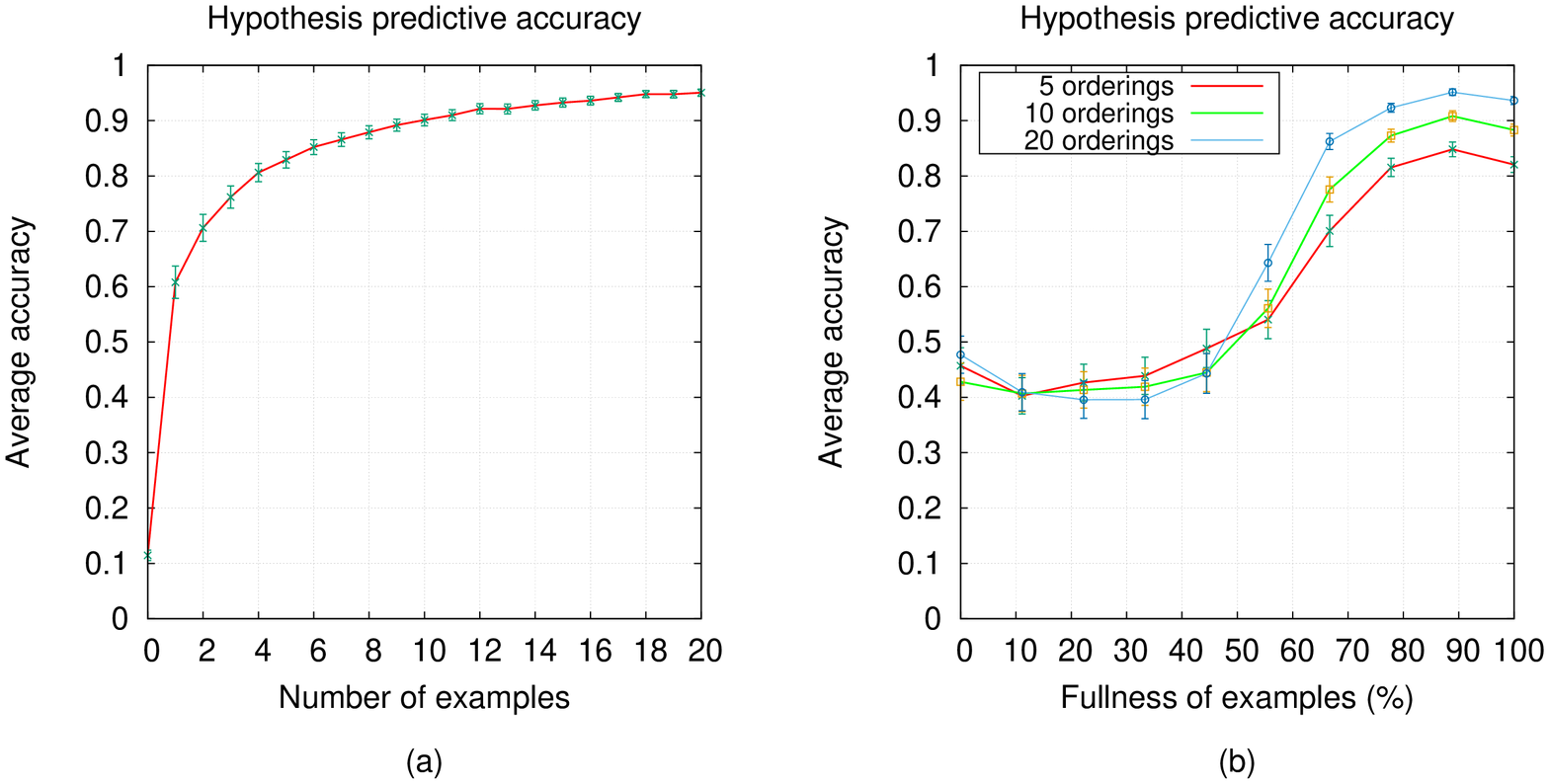}
  \caption{Accuracy with varying (a) numbers of examples; (b)
  fullness of examples} \label{fig:graph_no_egs}
\end{figure}

\vspace{-1.7mm}

For each experiment we randomly selected 100 hypotheses, each with between $1$
to $3$ weak constraints from $S_M$, omitting hypotheses that ranked all answer
sets equally.  The only atoms that vary in $B$ are the $\asp{assign}$'s. As
there are 9 different slots, there are $2^9$ answer sets of $B$ (and many more
partial interpretations which are extended by these answer sets). We say an
example partial interpretation is \emph{full} if it specifies the truth value
of all 9 $\asp{assign}$ atoms, otherwise we describe the \emph{fullness} as the
percentage of the 9 atoms which are specified. In both experiments (for each of
the 100 target hypotheses $H_T$), we generated ordered pairs of partial
interpretations $o = \langle e_1, e_2\rangle$ such that $o$ was bravely
respected. If $o$ was also cautiously respected, then it was given as a
cautious example (otherwise it was used as a brave example).  In our first
experiment we investigated the effect of varying the number of examples, and in
the second we investigated the effects of varying the fullness of the examples.

In both experiments, we tested our approach 20 times for each target hypothesis
$H_T$. Each time, we used $ILASP2$ to learn a hypothesis $H_L$ which covered all
examples. We then calculated the accuracy of $H_L$ in predicting the pairwise
ordering of answer sets in $B$ (for each pair of answer sets $A_1, A_2\in
AS(B)$ we tested whether $H_T$ and $H_L$ agreed on the preference between
them).

In our first experiment we investigated the effect of varying the number of
examples from $0$ to $20$. The examples were of random fullness, each with
between $5$ to $9$ $\asp{assign}$ atoms specified.
Figure~\ref{fig:graph_no_egs}(a) shows the average predictive accuracy. Each
point on the graph corresponds to 2000 learning tasks (100 target hypotheses
with 20 different sets of examples).  The error bars on the graph show the
standard error. The results show that our method achieves 90\% accuracy for
this experiment with around 10 or more random examples.

For our second experiment we again tested our approach on 100 randomly
generated hypotheses with 20 different sets of randomly generated examples.
This time, however, we have kept the number of examples fixed at $5$, $10$ and
$20$ and varied the fullness of the examples.  Results are shown in
Figure~\ref{fig:graph_no_egs}(b). The graph shows that examples are only useful
if they are more than 50\% full. One interesting point to note is that the peak
performance is with examples of around 90\% fullness. This is because cautious
ordering examples are actually more useful if they are less full (as there are
more pairs which extend them); however, orderings are less likely to be
cautiously respected when they are less full.

\begin{figure}[!Ht]
  \centering
  \includegraphics[trim=0cm 2.0cm 0cm 2cm, clip=true, width=\textwidth]{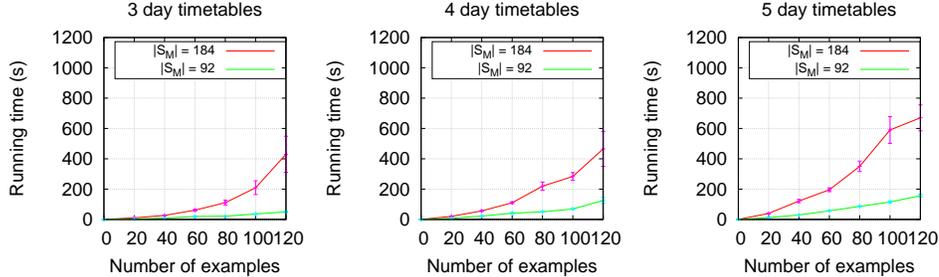}
  \caption{Average running time of ILASP2 with varying numbers of examples}
\label{fig:performance}
\vspace{-5mm}
\end{figure}

In our final experiment, we investigated the scalability of ILASP2 by
increasing both the number of days in our timetable and the number of examples.
Figure~\ref{fig:performance} shows the average running time for ILASP2 with 3,
4 and 5 day timetables (each with 3 slots) with up to 120 ordering examples.
The learning tasks are targeted at learning the hypothesis from
Example~\ref{eg:scheduleex}.  We randomly generated ordering examples, as in
the previous experiments with the slight difference that the fullness of the
examples was unrestricted.
As the hypothesis in these experiments does not use negative weights in either
of the weak constraints, we also tested the average running time with a search
space containing only positive weights. This means that $S_M$ contained 92 weak
constraints rather than the original 184.  These experiments show that the time
taken to solve an ILASP2 task is dependent not only on the number of examples,
but also on the size of the domain and the size of $S_M$.

\section{Related Work}
\label{sec:rel}

In~\cite{JELIAILASP} we showed that any of the learning tasks
in~\cite{Corapi2012,ray2009nonmonotonic,Sakama2009,Otero2001} could be
expressed by $ILP_{LAS}$ and computed by \emph{ILASP}.  As any
$ILP_{LAS}$ task can be (trivially) mapped into an $ILP_{LOAS}$ (i.e. 
$O^b = \emptyset$ and $O^c= \emptyset$), $ILP_{LOAS}$ inherits this property. None of the
previous learning tasks (including $ILP_{LAS}$), however, can construct
examples which incentivise the learning of a hypothesis containing a weak
constraint.  This is because they can only give examples of what should (or
shouldn't) be an answer set of $B\cup H$. In addition, $ILP_{LOAS}$ inherits
the capability of $ILP_{LAS}$ of supporting \emph{predicate invention},
allowing new concepts to be invented whilst learning.

The ILASP2 algorithm is an extension of the original ILASP algorithm
in~\cite{JELIAILASP}. It extends the concepts of positive and violating
hypothesis to cover learning weak constraints (which was not possible in
ILASP). For the simpler $ILP_{LAS}$ tasks, ILASP2 is more efficient than ILASP.
As discussed in section~\ref{sec:alg}, the original ILASP algorithm has some
scalability issues when there is a large number of violating hypotheses.  We
have shown in section~\ref{sec:experiments} that by eliminating violating
reasons rather than single violating hypotheses, ILASP2 can be much more
efficient.

Also related to our work are existing approaches for \emph{learning to rank}.
These use non logic-based machine learning techniques (e.g. neural
networks~\cite{geisler2001modeling}). Our approach shares the same advantages
as any ILP approach versus a non logic-based machine learning technique:
learned hypotheses are structured, human readable and can express relational
concepts such as minimising the instances of particular combinations of
predicates.  Existing background knowledge can be taken into account to capture
predefining concepts and the search can be steered towards particular types of
hypotheses using a language bias.  Furthermore, ILASP2 is also capable of
learning preferences with weights and priorities, meaning that more structured
and complex preferences can be learned.

An example of the use of an ILP system for learning constraints has been
recently presented in \cite{lallouet2010learning} where timetabling constraints
are learned from positive and negative examples. In this case the learned rules
are hard constraints (e.g., enforcing that a teacher is not in two places at
once). Examples of this kind are already computable by $ILP_{LAS}$, and so are
also computable by $ILP_{LOAS}$.

\vspace{-0.5mm}
\section{Conclusion and Future Work}
\label{sec:conc}

We have presented a new framework for ILP, Learning from Ordered Answer Sets,
which extends previous ILP systems in that it is able to learn weak constraints
and can be used to perform preference learning. The framework can represent
partial examples under a brave and a cautious semantics.  We have also put
forward a new algorithm, ILASP2, that can solve any $ILP_{LOAS}$ task for
optimal solutions. This algorithm extends previous work for solving the simpler
task $ILP_{LAS}$ and resolves some of the scalability issues associated with
the previous algorithm. Some scalability issues remain, especially when there
is a particularly large hypothesis space and future work will focus on
overcoming these. Current work also addresses extending the ILASP algorithm to
support noisy examples.

\pagebreak{}

\bibliographystyle{acmtrans}
\bibliography{paper}
\clearpage{}

\begin{appendix}

\section{The ILASP2 Meta Encoding}

We present here the ILASP2 meta encoding which is omitted from the main paper.
We first summarise some notation used in the encoding.

We will write $body^{+}(R)$ and $body^{-}(R)$ to refer to the positive and
negative (respectively) literals in the body of a rule $R$. Given a program
$P$, $weak(P)$ denotes the weak constraints in $P$ and $non\_weak(P)$ denotes
the set of rules in $P$ which are not weak constraints.

\begin{definition}
  For any ASP program $P$, predicate name $pred$ and term $\mathtt{term}$ we
  will write\break $reify(P, pred, term)$ to mean the program constructed by
  replacing every atom $\mathtt{a} \in P$ by $\mathtt{pred(a, term)}$.  We will
  use the same notation for sets of literals/partial interpretations, so for a
  set $S$: $reify(S, pred, term) = \lbrace \mathtt{pred(atom, term)} :
  \mathtt{atom} \in S \rbrace$.
\end{definition}

\begin{definition}
  For any ASP program $P$ and any atom $\mathtt{a}$, $append(P, a)$ is the
  program constructed by appending $\mathtt{a}$ to every rule in $P$.
\end{definition}

\begin{definition}
  Given any term $t$ and any positive example $e$, $cover(e,t)$ is the program:

  \begin{math}
    \begin{array}{l}
      \asp{cov(t)\codeif in\_as(e^{inc}_1, t), \ldots, in\_as(e^{inc}_n, t),
        \naf in\_as(e^{exc}_1, t),\ldots, \naf in\_as(e^{exc}_m, t)\ruleend}\\
      \asp{\codeif \naf cov(t)\ruleend}
    \end{array}
  \end{math}
\end{definition}

The previous three definitions can be used in combination to test whether a
program has an answer sets which extend given partial interpretations.

\begin{example}

Consider the program $P = \left\{\begin{array}{l}
  \asp{p \codeif \naf q\ruleend}\\
  \asp{q \codeif \naf p\ruleend}
\end{array}\right\}$ and the partial interpretations $I_1 = \langle \lbrace
p\rbrace, \emptyset\rangle$ and $I_2 = \langle \emptyset, \lbrace
p\rbrace\rangle$.

The program $Q = append(reify(P, in\_as, X), as(X))\cup \lbrace \asp{as(as1)},
\asp{as(as2)}\rbrace \cup\break cover(I_1, as1)\cup cover(I_2, as2)$ has the grounding:

{\small
\begin{verbatim}
in_as(p, as1) :- not in_as(q, as1), as(as1).
in_as(q, as1) :- not in_as(p, as1), as(as1).
in_as(p, as2) :- not in_as(q, as2), as(as2).
in_as(q, as2) :- not in_as(p, as2), as(as2).

as(as1).
as(as2).

cov(as1) :- in_as(p, as1).
:- not cov(as1).

cov(as2) :- not in_as(p, as2).
:- not cov(as2).
\end{verbatim}
}

Without the two constraints, $Q$ would have 4 answer sets (the combinations of
$\asp{as1}$ and $\asp{as2}$ corresponding to the two answer sets of $P$). With
the two constraints, the answer set represented by $\asp{as1}$ must extend
$I_1$, and the answer set represented by $\asp{as2}$ must extend $I_2$. Hence,
there is only one answer set of $Q$, $\lbrace \asp{as(as1)}$, $\asp{as(as2)}$,
$\asp{in\_as(p, as1)}$, $\asp{in\_as(q, as2)}$, $\asp{cov(as1)}$,
$\asp{cov(as2)}\rbrace$.

Note that the answer sets of $P$ which extend $I_1$ and $I_2$ ($\lbrace
\asp{p}\rbrace$ and $\lbrace \asp{q}\rbrace$) can be extracted from the
$in\_as$ atoms in this answer set of $Q$. If there were multiple answer sets of
$P$ extending one or more of the partial interpretations, then there would be
multiple answer sets, representing all possible combinations such that the
constraints are met.

\end{example}

\begin{definition}
  Let $p_1$ and $p_2$ be distinct predicate names and $t$ be a term. Given $R$,
a weak constraint $\asp{:\sim b_1,\ldots,b_m,}$ $\asp{\textrm{not
}c_1,\ldots,\textrm{not }c_l\ruleend[wt@lev, t_1,\ldots, t_n]}$,
$meta_{weak}(R, p_1, p_2, t)$ is the rule:

\noindent
$\asp{w(wt, lev, args(t_1,\ldots,t_n), t) \codeif p_2(X), p_1(b_1,t),
\ldots, p_1(b_m, t),}\asp{\naf p_1(c_1,t), \ldots , \naf p_1(c_l,t)\ruleend}$

\noindent
  For a set of weak constraints $W$, $meta_{weak}(W, p_1, p_2, t)$ is the set
\break $\lbrace meta_{weak}(R, p_1, p_2, t) \mid R \in W\rbrace$.

\end{definition}

\begin{example}\label{eg:weak_rep}

Consider the program $P$ containing the two weak constraints:

{\small
\begin{verbatim}
:~ p(V).[1@2, V]
:~ q(V).[2@1, V]
\end{verbatim}
}

$meta_{weak}(P, in\_as, as, X)$ is the program:

{\small
\begin{verbatim}
w(1, 2, args(V), X) :- as(X), in_as(p(V), X).
w(2, 1, args(V), X) :- as(X), in_as(q(V), X).
\end{verbatim}
}

Note that for any program $P$, if we reify an interpretation $I = \lbrace
\asp{a_1}, \ldots, \asp{a_n}\rbrace$ as $\lbrace \asp{in\_as(a_1, id)}, \ldots,
\asp{in\_as(a_n, id)}\rbrace$ (the set $reify(I, in\_as, id)$) then the atoms\break
$\asp{w(wt, l, args(t_1,\ldots, t_m), id)}$ in the (unique) answer set of\break
$meta_{weak}(weak(P), in\_as, as, X)\cup reify(I, in\_as, id)\cup\lbrace
\asp{as(id)}\rbrace$ correspond exactly to the elements $(wt, l, t_1,\ldots
t_m)$ of $weak(P, I)$.

For example, consider the interpretation $I = \lbrace p(1), p(2), q(1)\rbrace$.
The unique answer set of $meta_{weak}(weak(P), in\_as, as, X)\cup reify(I, in\_as,
id)\cup\lbrace \asp{as(id)}\rbrace$ is $\lbrace \asp{as(id)},$\break $\asp{in\_as(p(1),
id)},$ $\asp{in\_as(p(2), id)},$ $\asp{in\_as(q(1), id)},$ $\asp{w(1, 2, args(1), id)},$
$\asp{w(1, 2, args(2), id)},$\break $\asp{w(2, 1, args(1), id)}\rbrace$. In this case,
$weak(P, I) = \lbrace (1, 2, 1), (1, 2, 2), (2, 1, 1)\rbrace$.

\end{example}

Now that we have defined the predicate $w$ to represent $weak(P, A)$ for each
answer set $A$, we can use some additional rules to determine, given two
interpretations, whether one dominates another.

\begin{definition}

  Given any two terms $t1$ and $t2$, $dominates(t1,t2)$ is the program:

\begin{math}
\left\{\begin{array}{l}
\asp{dom\_lv(t1,t2,L) \codeif lv(L), \texttt{\#sum}\lbrace w(W,L,A,t1)=W, w(W,L,A,t2)=-W\rbrace < 0\ruleend}\\
\asp{non\_dom\_lv(t1,t2,L) \codeif lv(L), \texttt{\#sum}\lbrace w(W,L,A,t2)=W, w(W,L,A,t1)=-W\rbrace < 0\ruleend}\\
\asp{non\_bef(t1,t2,L) \codeif lv(L), lv(L2), L < L2, non\_dom\_lv(t1,t2,L2)\ruleend}\\
\asp{dom(t1,t2) \codeif dom\_lv(t1,t2,L), \naf non\_bef(t1,t2,L)\ruleend}\\
\end{array}\right\}
\end{math}

\end{definition}

The intuition is that $\asp{dom(id1, id2)}$ (where $\asp{id1}$ and $\asp{id2}$
represent two answer sets $A_1$ and $A_2$) should be true if and only if $A_1$
dominates $A_2$. This is dependent on the atoms $\asp{dom\_lv(id1, id2, l)}$,
which for each level $l$, should be true if and only if $P_{A_1}^l <
P_{A_2}^l$; $\asp{non\_dom\_lv(id1, id2, l)}$, which for each level $l$, should
be true if and only if $P_{A_2}^l < P_{A_1}^l$; and finally,
$\asp{non\_bef(id1, id2, l)}$, which for each level $l$, should be true if and
only if there is an $l_2 > l$ such that $P_{A_2}^{l_2} < P_{A_1}^{l_2}$.

\begin{example}

Consider again the program $P$ and interpretation $I$ from
example~\ref{eg:weak_rep}. Consider also an additional interpretation $I' =
\lbrace p(1), p(2), p(3)\rbrace$. $weak(P, I') = \lbrace (1, 2, 1), (1, 2, 2),$
$(1, 2, 3)\rbrace$.

The unique answer set of $meta_{weak}(weak(P), in\_as, as, X) \cup reify(I,
in\_as, id1) \cup reify(I', in\_as, id2)\cup \lbrace \asp{as(id1)},
\asp{as(id2)}, \asp{lv(1)}, \asp{lv(2)} \rbrace \cup dominates(id1, id2)$
contains $\asp{dom\_lv(id1, id2, 2)}$, because $I$ dominates $I'$ at level 2
(i.e.\ $P_I^{2} < P_{I'}^{2}$); contains $\asp{non\_dom\_lv(id1, id2, 1)}$,
because $I'$ dominates $I$ at level 1; does not contain any $non\_bef$ atoms,
because the only level at which $I'$ dominates $I$ is 1, which is not evaluated
``before'' any other level (it is the lowest level in the program); and
finally, does contain $\asp{dom(id1, id2)}$ because $I$ dominates $I'$ at level
2 and there is no level ``before'' (higher than) level 2 at which $I'$
dominates $I$. The presence of $\asp{dom(id1, id2)}$ in the answer set
indicates that $I$ dominates $I'$.

Similarly, the unique answer set of $meta_{weak}(weak(P), in\_as, as, X) \cup\break
reify(I, in\_as, id1) \cup reify(I', in\_as, id2)\cup \lbrace \asp{as(id1)},
\asp{as(id2)}, \asp{lv(1)}, \asp{lv(2)} \rbrace \cup\break dominates(id2, id1)$
contains $\asp{dom\_lv(id2, id1, 1)}$, because $I'$ dominates $I$ at level 1;
contains $\asp{non\_dom\_lv(id2, id1, 2)}$, because $I$ dominates $I'$ at level
2; contains\break $\asp{non\_bef(id2, id1, 1)}$ as $I$ dominates $I'$ at level 2,
which is evaluated ``before'' level 1; and finally, does not contain
$\asp{dom(id2, id1)}$ because there is no level $l$ in the program such that
$I'$ dominates $I$ at $l$ and $I$ does not dominate $I'$ at any higher level.

\end{example}

\subsection{Encoding the search for positive hypotheses: $T_{meta}$}

We now use the components described in the previous section to define a program
$T_{meta}$ whose answer sets correspond to the positive solutions of an
$ILP_{LOAS}$ task $T$.

\begin{definition}

Let $T$ be the $ILP_{LOAS}$ task $\langle B, S_M, E^{+}, E^{-}, O^{b},
O^{c}\rangle$.  Then $T_{meta} = meta(B)\cup meta(S_M) \cup meta(E^{+}) \cup
meta(E^{-}) \cup meta(O^{b}) \cup meta(O^{c})$ where each meta component is as
follows:

  \begin{itemize}
    \item $meta(B) = append(reify(non\_weak(B), in\_as, X),
      as(X))\\\mbox{\hspace{30mm}}\cup meta_{weak}(weak(B), in\_as, as, X)$.
    \item  $meta(S_M)=\\\mbox{\hspace{1mm}}\lbrace append(append(reify(R, in\_as, X), as(X)),
      in\_h(R_{id})) \mid R \in non\_weak(S_M)\rbrace\\
      \tab\cup\lbrace append(W, in\_h(W_{id})) \mid W \in
      meta_{weak}(weak(S_M), in\_as, as, X)\rbrace$\\
      $\tab\cup \lbrace \asp{:\sim
      in\_h(R_{id})\ruleend[2*|R|@0,R_{id}]}\mid R \in
      S_M\rbrace$\\
      $\tab\cup \lbrace\;\;\asp{ \lbrace in\_h(R_{id}) : R \in S_M
      \rbrace\ruleend}\;\;\rbrace$
    \item
      \begin{math}
        meta(E^{+}) = \left\{\begin{array}{l}
          cover(e, e_{id})\\
          \asp{as(e_{id})\ruleend}
        \end{array}\middle|
          \langle e^{inc}, e^{exc}\rangle \in E^{+}
        \right\}
      \end{math}
    \item
      \begin{math}
        meta(E^{-}) =\left\{\begin{array}{l}
          \asp{v\_i\codeif in\_as(e^{inc}_1, n), \ldots,
            in\_as(e^{inc}_n, n),}\\\tab
            \asp{\naf in\_as(e^{exc}_1, n),\ldots,}\\\tab
            \asp{\naf
            in\_as(e^{exc}_m, n)\ruleend}\\
          \asp{as(n)\ruleend}
        \end{array}\middle|
          \langle e^{inc}, e^{exc}\rangle \in E^{-}
        \right\}
        \\\tab\tab\tab\tab\tab\cup\left\{\begin{array}{l}
          \asp{violating\codeif v\_i\ruleend}\\
          \asp{:\sim\naf violating\ruleend[1@0]}
        \end{array}\right\}
      \end{math}
    \item
      \begin{math}
        meta(O^{b})=\left\{\begin{array}{l}
          \asp{as(o_{id1})\ruleend}\quad\quad\asp{as(o_{id2})\ruleend}\\
          cover(e^1, o_{id1})\\
          cover(e^2, o_{id2})\\
          dominates(o_{id1},o_{id2})\\
          \asp{\codeif \naf dom(o_{id1}, o_{id2})\ruleend}\\
        \end{array}\middle|
        \begin{array}{c}
          o=\langle e_1, e_2\rangle\in O^{b}
        \end{array}
        \right\}
        \cup\left\{ \asp{lv(l)\ruleend} \mid l \in L\right\}
      \end{math}
    \item
      \begin{math}
        meta(O^{c})=\left\{\begin{array}{l}
          dominates(e_{1},e_{2})\\
          \asp{v\_p(e^1_{id},e^2_{id})\codeif}\\\asp{\tab \naf dom(e_{1},
          e_{2})\ruleend}\\
        \end{array}\middle|
        \begin{array}{c}
          \langle e_1, e_2\rangle\in O^{c}
        \end{array}
        \right\}\cup\left\{\begin{array}{l}
          \asp{v\_p\codeif v\_p(T1,T2)\ruleend}\\
          \asp{violating\codeif v\_p\ruleend}\\
        \end{array}\right\}
      \end{math}
  \end{itemize}
\end{definition}

The intuition is that the $\asp{in\_h}$ atoms correspond to the rules in the
hypothesis. Each rule $R\in S_M$ has a unique identifier $R_{id}$ and if
$\asp{in\_h(R_{id})}$ is true then $R$ is considered to be part of the
hypothesis $H$. These $\asp{in\_h}$ atoms have been added to the bodies of the
rules in the meta encoding so that a rule $R\in S_M$ only has an effect if it
is part of $H$.

Each of the terms $t$ for which there is a fact $\asp{as(t)}$ represents an
answer set of $B\cup H$. As in the previous section the $cover$ program
$cover(I, t)$ is used to enforce that some of these answer sets extend
particular partial interpretations.

There is one $\asp{as(t)}$ atom for each positive example $e$. The $cover$
program is used to ensure that the corresponding answer set does extend $e$.
There are two $\asp{as(t)}$ atoms for each brave ordering $\langle e_1,
e_2\rangle$. Two instances of the $cover$ program are used to ensure that the
first answer set extends $e_1$ and the second answer set extends $e_2$. We also
use the $dominates$ program from the previous section and a constraint to
ensure that the first answer set dominates the second (hence the ordering is
bravely respected).

For the negative examples and the cautious orderings, the aim is to generate
$\asp{violating}$ in at least one answer set of the meta encoding corresponding
to $H$, if $H$ is indeed a violating solution (generating $\asp{v\_i}$ if $H$
does not cover a negative example and some instance of $\asp{v\_p}$ if it does
not respect a cautious ordering).

Firstly, for the negative examples, we have an extra fact $\asp{as(n)}$. As we
have no constraints on the answer set of $B\cup H$ which this can correspond
to, the intuition is that there is one answer set of the meta encoding for each
answer set of $B\cup H$. For each negative example $e^{-}$ there is a rule for
$\asp{v\_i}$ which will generate $\asp{v\_i}$ if the answer set corresponding
to $\asp{as(n)}$ extends $e^{-}$.

For the cautious orderings we use a similar approach. For any cautious ordering
$\langle e_1, e_2\rangle$, as $e_1$ and $e_2$ are positive examples, there are
already two $\asp{as(t)}$ atoms which represent answer sets extending each of
these interpretations; in fact, there will be one answer set of the meta
encoding for each possible pair of answer sets of $B\cup H$ which extend these
interpretations. Therefore, by using the $dominates$ program, and generating a
$\asp{v\_p}$ atom if the answer set of $B\cup H$ corresponding to the first
$\asp{as(t)}$ atom does not dominate the answer set of $B\cup H$ corresponding
to the second $\asp{as(t)}$ atom in any answer set of the meta encoding, we
ensure that $\asp{violating}$ will be true in at least one answer set of the
meta encoding which corresponds to $H$.

\begin{example}\label{eg:tenc}

\begin{multicols}{2}

Consider the learning task:

\noindent
\begin{math}
B = \left\{\begin{array}{l}
\asp{p(V) \codeif r(V), \naf q(V)\ruleend}\\
\asp{q(V) \codeif r(V), \naf p(V)\ruleend}\\
\asp{r(1)\ruleend}\quad\quad\asp{r(2)\ruleend}\\
\asp{a \codeif \naf b\ruleend}\\
\asp{b \codeif \naf a\ruleend}\\
\end{array}\right\}
\end{math}

\noindent
\begin{math}
S_M = \left\{\begin{array}{l}
\asp{q(1)\ruleend}\\
\asp{:\sim q(V)\ruleend[1@1, V, r2]}\\
\asp{:\sim b\ruleend[1@1, b, r3]}
\end{array}
\right\}
\end{math}

\noindent
\begin{math}
  E^{+} = \left\{
    \begin{array}{l}
      \langle\lbrace \asp{p(2)}\rbrace, \emptyset\rangle,\\
      \langle\emptyset, \lbrace \asp{p(2)} \rbrace\rangle,\\
      \langle\lbrace \asp{a}\rbrace, \lbrace \asp{b}\rbrace\rangle,\\
      \langle\emptyset, \lbrace \asp{a} \rbrace\rangle\\
    \end{array}
    \right\}
  \end{math}

\noindent
\begin{math}
  E^{-} = \left\{
    \begin{array}{l}
      \langle\lbrace \asp{p(1)}\rbrace, \emptyset\rangle\\
    \end{array}
    \right\}
  \end{math}

\noindent
\begin{math}
  O^{b} = \left\{
    \begin{array}{l}
      \langle e^{+}_{3}, e^{+}_{4} \rangle\\
    \end{array}
    \right\}
  \end{math}

\noindent
\begin{math}
  O^{c} = \left\{
    \begin{array}{l}
      \langle e^{+}_{1}, e^{+}_{2} \rangle\\
    \end{array}
    \right\}
  \end{math}
\end{multicols}

Figure~\ref{fig:tmeta} shows $T_{meta}$. There are two optimal positive
hypotheses (one containing each of the two weak constraints).

The positive hypothesis $\asp{:\sim b.[1@1,b,r3]}$ has a violating
interpretation $\lbrace \asp{p(1)}, \asp{q(2)},$\break $\asp{r(1)}, \asp{r(2)},
\asp{a}\rbrace$. This corresponds to the following answer set of $T_{meta}$:

{\small
\begin{verbatim}
{ as(1), as(2), as(3), as(4), as(n), as(5), as(6), lv(1), in_as(r(1),1),
  in_as(r(1),2), in_as(r(1),3), in_as(r(1),4), in_as(r(1),n), in_as(r(1),5),
  in_as(r(1),6), in_as(r(2),1), in_as(r(2),2), in_as(r(2),3), in_as(r(2),4),
  in_as(r(2),n), in_as(r(2),5), in_as(r(2),6), in_as(q(1),3), in_as(q(1),4),
  in_as(q(1),6), in_as(p(1),1), in_as(p(1),2), in_as(p(1),n), in_as(p(1),5),
  in_as(p(2),1), in_as(q(2),2), in_as(q(2),3), in_as(q(2),4), in_as(q(2),n),
  in_as(q(2),5), in_as(q(2),6), in_as(a,1), in_as(a,2), in_as(a,3),
  in_as(b,4), in_as(a,n), in_as(a,5), in_as(b,6), in_h(r2),
  w(1,1,args(2,r2),2), w(1,1,args(1,r2),3), w(1,1,args(2,r2),3),
  w(1,1,args(1,r2),4), w(1,1,args(2,r2),4), w(1,1,args(2,r2),n),
  w(1,1,args(2,r2),5), w(1,1,args(1,r2),6), w(1,1,args(2,r2),6), cov(1),
  cov(2), cov(3), cov(4), v_i, violating, dom_lv(1,2,1), dom(1,2), cov(5),
  cov(6), dom_lv(5,6,1), dom(5,6) }
\end{verbatim}
}

Note that the violating interpretation can be extracted from the
$\asp{in\_as(\_, n)}$ atoms and the hypothesis can be extracted from the
$\asp{in\_h}$ atoms.

Similarly, the violating pair $\langle \lbrace \asp{p(2)}, \asp{q(1)},
\asp{r(1)}, \asp{r(2)}, \asp{a}\rbrace, \lbrace \asp{p(1)}, \asp{q(2)},
\asp{r(1)}, \asp{r(2)}, \asp{a}\rbrace\rangle$ can be extracted from the answer
set:

{\small
\begin{verbatim}
{ as(1), as(2), as(3), as(4), as(n), as(5), as(6), lv(1), in_as(r(1),1),
  in_as(r(1),2), in_as(r(1),3), in_as(r(1),4), in_as(r(1),n), in_as(r(1),5),
  in_as(r(1),6), in_as(r(2),1), in_as(r(2),2), in_as(r(2),3), in_as(r(2),4),
  in_as(r(2),n), in_as(r(2),5), in_as(r(2),6), in_as(q(1),1), in_as(q(1),4),
  in_as(q(1),6), in_as(p(1),2), in_as(p(1),3), in_as(p(1),n), in_as(p(1),5),
  in_as(p(2),1), in_as(q(2),2), in_as(q(2),3), in_as(q(2),4), in_as(q(2),n),
  in_as(q(2),5), in_as(q(2),6), in_as(a,1), in_as(a,2), in_as(a,3),
  in_as(b,4), in_as(a,n), in_as(a,5), in_as(b,6), w(1,1,args(1,r2),1),
  in_h(r2), w(1,1,args(2,r2),2), w(1,1,args(2,r2),3), w(1,1,args(1,r2),4),
  w(1,1,args(2,r2),4), w(1,1,args(2,r2),n), w(1,1,args(2,r2),5),
  w(1,1,args(1,r2),6), w(1,1,args(2,r2),6), cov(1), cov(2), cov(3), cov(4),
  v_i, violating, v_p(1,2), v_p, cov(5), cov(6), dom_lv(5,6,1), dom(5,6) }
\end{verbatim}
}

\begin{figure*}[ht!]
  \begin{multicols}{2}
{\small
\begin{verbatim}

% meta(B)
in_as(p(V),X) :- in_as(r(V),X),
  not in_as(q(V),X), as(X).
in_as(q(V),X) :- in_as(r(V),X),
  not in_as(p(V),X), as(X).
in_as(r(1),X) :- as(X).
in_as(r(2),X) :- as(X).
in_as(a,X) :- not in_as(b,X), as(X).
in_as(b,X) :- not in_as(a,X), as(X).

% meta(S_M)
in_as(q(1),X) :- as(X), in_h(r1).
w(1,1,args(V,r2),X) :- in_as(q(V),X),
  as(X), in_h(r2).
w(1,1,args(b,r3),X) :- in_as(b,X),
  as(X), in_h(r3).
0 {in_h(r1), in_h(r2), in_h(r3)} 2.
:~ in_h(r1).[2@0,r1]
:~ in_h(r2).[2@0,r2]
:~ in_h(r3).[2@0,r3]

% meta(E^+)
as(1).
as(2).
as(3).
as(4).
cov(1) :- in_as(p(2),1).
cov(2) :- not in_as(p(2),2).
cov(3) :- in_as(a,3), not in_as(b,3).
cov(4) :- not in_as(a,4).
:- not cov(1).
:- not cov(2).
:- not cov(3).
:- not cov(4).



% meta(E^-)
v_i :- in_as(p(1),n).
as(n).
violating :- v_i.
:~ not violating.[1@0, violating]

% meta(O^b)
as(5).
as(6).
cov(5) :- in_as(a,5), not in_as(b,5).
cov(6) :- not in_as(a,6).
:- not cov(5).
:- not cov(6).
dom_lv(5,6,L) :- lv(L),
  #sum{w(W,L,A,5)=W, w(W,L,A,6)=-W} < 0.
wrong_dom_lv(5,6,L) :- lv(L),
  #sum{w(W,L,A,6)=W, w(W,L,A,5)=-W} < 0.
wrong_bef(5,6,L) :- lv(L), L < L2,
  wrong_dom_lv(5,6,L2).
dom(5,6) :- dom_lv(5,6,L),
  not wrong_bef(5,6,L).
:- not dom(5,6).
lv(1).

% meta(O^c)
dom_lv(1,2,L) :- lv(L),
  #sum{w(W,L,A,1)=W, w(W,L,A,2)=-W} < 0.
wrong_dom_lv(1,2,L) :- lv(L),
  #sum{w(W,L,A,2)=W, w(W,L,A,1)=-W} < 0.
wrong_bef(1,2,L) :- lv(L), L < L2,
  wrong_dom_lv(1,2,L2).
dom(1,2) :- dom_lv(1,2,L),
  not wrong_bef(1,2,L).
v_p(1,2) :- not dom(1,2).
v_p :- v_p(X,Y).
violating :- v_p.
\end{verbatim}
}
\end{multicols}
\caption{An example of $T_{meta}$.\label{fig:tmeta}}
\end{figure*}
\end{example}

\pagebreak{}
\subsection{Encoding classes of violating hypotheses: $VR_{meta}$}

The previous section set out how to construct a meta level ASP program
$T_{meta}$ whose answer sets correspond to the positive hypotheses. It is also
able to identify violating hypotheses, but has no way of eliminating them (as
only a subset of the meta level answer sets corresponding to a violating
hypothesis will identify it as violating, so eliminating these answer sets does
not necessarily eliminate the hypothesis).

We therefore present a new meta level program which, combined with $T_{meta}$,
eliminates any hypothesis which is violating for a given set of violating
reasons. As violating reasons are full answer sets (or pair of full answer
sets), we do not have to generate answer sets. It is enough to check whether
the answer sets we have as part of the violating reasons are still answer sets
of $B\cup H$. The standard way to do this is check whether these answer sets
are the minimal model of the reduct of $B\cup H$ with respect to this answer
set. The next two definitions define a meta level program which, given an
interpretation, computes the minimal model of the reduct of a program with
respect to that interpretation. This reduct construction is close to the
simplified reduct for choice rules from~\cite{simpleReduct}.

\begin{definition}

  Given any choice rule $R=\;\asp{l \lbrace h_1,\ldots, h_n \rbrace u\codeif
  body}$, $reductify(R)$ is the program:

\noindent
  \begin{math}
    \tab\left\{\begin{array}{c}
        \asp{mmr(h_1,X) \codeif reify(body^{+}, mmr, X), reify(body^{-},
        \naf in\_vs, X),}\\\tab\tab \asp{l \lbrace in\_vs(h_1, X), \ldots, in\_vs(h_n, X) \rbrace u,
        in\_vs(h_1,X).}\\
          \ldots\\
        \asp{mmr(h_n,X) \codeif reify(body^{+}, mmr, X), reify(body^{-},
        \naf in\_vs, X),}\\\tab\tab \asp{l \lbrace in\_vs(h_1, X), \ldots, in\_vs(h_n, X) \rbrace u,
        in\_vs(h_n,X).}\\
        \asp{mmr(\bot, X) \codeif reify(body^{+}, mmr, X), reify(body^{-},
        \naf in\_vs, X),}\\\tab\tab \asp{u+1 \lbrace in\_vs(h_1, X), \ldots, in\_vs(h_n, X) \rbrace.}\\
        \asp{mmr(\bot, X) \codeif reify(body^{+}, mmr, X), reify(body^{-},
        \naf in\_vs, X),}\\\tab\tab \asp{ \lbrace in\_vs(h_1, X), \ldots, in\_vs(h_n, X) \rbrace l-1.}
      \end{array}
    \right\}
  \end{math}
\end{definition}

\begin{definition}
  Let $P$ be an ASP program such that $P_1$ is the set of normal rules in $P$,
$P_2$ is the set of constraints in $P$ and $P_3$ is the set of choice rules in
$P$.

\noindent
\begin{math}
  reductify(P) = \left\{\begin{array}{l}
    \asp{mmr(head(R), X)\codeif reify(body^{+}(R),mmr,X),}\\
    \tab\asp{reify(body^{-}(R), \naf in\_vs, X), vs(X)\ruleend}
  \end{array} \middle| R\in P_1\right\}\\
  \tab\tab\tab\tab\tab\cup \left\{\begin{array}{l}
    \asp{mmr(\bot, X)\codeif reify(body^{+}(R),mmr,X),}\\
    \tab\asp{reify(body^{-}(R), \naf in\_vs, X), vs(X)\ruleend}
  \end{array} \middle| R\in P_2\right\}\\
  \tab\tab\tab\tab\tab\cup \lbrace reductify(R) \mid R\in P_3\rbrace.
\end{math}

\end{definition}

\begin{example}

Consider the program $P = \left\{\begin{array}{c}
  \asp{p \codeif \naf q\ruleend}\\
  \asp{q \codeif \naf p\ruleend}
\end{array}\right\}$

$reductify(P) = \left\{\begin{array}{c}
  \asp{mmr(p, X) \codeif \naf in\_vs(q, X), vs(X)\ruleend}\\
  \asp{mmr(q, X) \codeif \naf in\_vs(p, X), vs(X)\ruleend}\\
\end{array}\right\}$

We can check whether $\lbrace \asp{p}\rbrace$ is an answer set by combining
$reductify(P)$ with $\lbrace \asp{vs(vs1)}, \asp{in\_vs(p, vs1)}\rbrace$.  The
answer set of this program is $\lbrace \asp{vs(vs1)}, \asp{in\_vs(p, vs1)},\break
\asp{mmr(p, vs1)}\rbrace$, from which the minimal model, $\lbrace
\asp{p}\rbrace$, can be extracted. This shows that $\lbrace \asp{p}\rbrace$ is
indeed an answer set of $P$.

\end{example}

\begin{definition}
  Let $T$ be the $ILP_{LOAS}$ task $\langle B, S_M, E^{+}, E^{-}, O^{b},
  O^{c}\rangle$ and $VR$ be the set of violating reasons $VI\cup VP$, where
  $VI$ are violating interpretations and $VP$ are violating pairs.

  $VR_{meta}(T)$ is the program $meta(VI)\cup meta(VP)\cup meta(Aux)$ where the
  meta components are defined as follows:

\begin{itemize}
\item
  \begin{math}
      meta(VI) = \left\{\begin{array}{l}
          reify(I,in\_vs, I_{id})\\
          \asp{\codeif \naf nas(I_{id})\ruleend}\\
          \asp{vs(I_{id})\ruleend}
        \end{array}\middle|
        I \in VI
      \right\}
  \end{math}
\item
  \begin{math}
    meta(VP) = \left\{\begin{array}{l}
        dominates(vp_{id1},vp_{id2})\\
        reify(I_1, in\_vs, vp_{id1})\\
        reify(I_2, in\_vs, vp_{id2})\\
        \asp{vs(vp_{id1})\ruleend}\\
        \asp{vs(vp_{id2})\ruleend}\\
        \codeif \asp{\naf nas(vp_{id1}),\naf  nas(vp_{id2}),}\\\tab
          \asp{\naf dom(vp_{id1},vp_{id2})\ruleend}
    \end{array}\middle|
      vp = \langle I_1, I_2\rangle \in VP
    \right\}
  \end{math}
\item
  \begin{math}
    meta(Aux) =\\
    \left\{\begin{array}{l}
      reductify(B)
      \\\mbox{\hspace{2mm}} \cup\left\{\begin{array}{l}
        \asp{nas(X) \codeif in\_vs(ATOM, X), \naf mmr(ATOM, X)\ruleend}\\
        \asp{nas(X) \codeif \naf in\_vs(ATOM, X), mmr(ATOM, X)\ruleend}
      \end{array}\right\}
      \\\mbox{\hspace{2mm}}\cup
      \lbrace append(reductify(R), in\_hyp(R_{id})) \mid R \in
      non\_weak(S_M)\rbrace
      \\\mbox{\hspace{2mm}}\cup
      \lbrace append(meta_{weak}(W, in\_vs, vs, X), in\_hyp(W_{id})) \mid W \in weak(S_M)\rbrace
      \\\mbox{\hspace{2mm}}\cup
      \lbrace meta_{weak}(W, in\_vs, vs, X) \mid W \in weak(B)\rbrace
      \\\mbox{\hspace{2mm}}\cup\lbrace \asp{lv(l)\ruleend}\mid l \in L\rbrace
    \end{array}
    \right\}
  \end{math}
\end{itemize}
\end{definition}

This meta encoding uses the reductify program to check whether the various
interpretations in each violating reason is still an answer set of $B\cup H$.
There is a constraint for each of the violating interpretation, ensuring that
it is no longer an answer set of $B\cup H$. Similarly, there is a constraint
for each violating pair that says, if both interpretations are still answer
sets of $B\cup H$, then the first must dominate the second. This is checked by
using the $dominates$ program as before (the weak constraints are also
translated as before).

\begin{example}

Recall $B$, $S_M$, $E^{+}$, $E^{-}$, $O^{b}$ and $O^{c}$ from
example~\ref{eg:tenc} and let $VI$ be the set containing the violating
interpretation $\lbrace \asp{p(1), p(2), r(1), r(2), a}\rbrace$, $VP$ the set
containing the violating pair $\langle \lbrace \asp{p(2), q(1), r(1), r(2),
a}\rbrace, \lbrace\asp{q(1), q(2), r(1), r(2), a}\rbrace\rangle$ and let $VR$
be the set of violating reasons $VI\cup VP$. Then figure~\ref{fig:vmeta} shows
$VR_{meta}(T)$.

Now that we have ruled out any hypothesis with these violating reasons, one
optimal answer set of this program is:

{\small
\begin{verbatim}
{ as(1), as(2), as(3), as(4), as(n), as(5), as(6), lv(1), in_vs(p(1),v1),
  in_vs(p(2),v1), in_vs(r(1),v1), in_vs(r(2),v1), in_vs(a,v1), vs(v1),
  in_vs(p(2),v2), in_vs(q(1),v2), in_vs(r(1),v2), in_vs(r(2),v2), in_vs(a,v2),
  vs(v2), in_vs(q(1),v3), in_vs(q(2),v3), in_vs(r(1),v3), in_vs(r(2),v3),
  in_vs(a,v3), vs(v3), in_as(r(1),1), in_as(r(1),2), in_as(r(1),3),
  in_as(r(1),4), in_as(r(1),n), in_as(r(1),5), in_as(r(1),6), in_as(r(2),1),
  in_as(r(2),2), in_as(r(2),3), in_as(r(2),4), in_as(r(2),n), in_as(r(2),5),
  in_as(r(2),6), in_as(q(1),1), in_h(r1), in_as(q(1),2), in_as(q(1),3),
  in_as(q(1),4), in_as(q(1),n), in_as(q(1),5), in_as(q(1),6), in_as(p(2),1),
  in_as(q(2),2), in_as(p(2),3), in_as(q(2),4), in_as(p(2),n), in_as(p(2),5),
  in_as(q(2),6), in_as(a,1), in_as(a,2), in_as(a,3), in_as(b,4), in_as(a,n),
  in_as(a,5) in_as(b,6), w(1,1,args(1,r2),1), in_h(r2), w(1,1,args(1,r2),2),
  w(1,1,args(2,r2),2), w(1,1,args(1,r2),3), w(1,1,args(1,r2),4),
  w(1,1,args(2,r2),4), w(1,1,args(1,r2),n), w(1,1,args(1,r2),5),
  w(1,1,args(1,r2),6), w(1,1,args(2,r2),6), cov(1), cov(2), cov(3), cov(4),
  w(1,1,ts(1),v2), w(1,1,ts(1),v3), w(1,1,ts(2),v3), dom_lv(1,2,1), dom(1,2),
  cov(5), cov(6), dom_lv(5,6,1), dom(5,6), mmr(r(1),v1), mmr(r(1),v2),
  mmr(r(1),v3), mmr(r(2),v1), mmr(r(2),v2), mmr(r(2),v3), mmr(p(1),v1),
  mmr(p(2),v1), mmr(p(2),v2), mmr(q(1),v2), mmr(q(1),v3), mmr(q(2),v3),
  mmr(a,v1), mmr(a,v2), mmr(a,v3), mmr(q(1),v1), nas(v1), dom_lv(v2,v3,1),
  dom(v2,v3) }
\end{verbatim}
}

This answer set corresponds to the hypothesis:

{\small
\begin{verbatim}
q(1).
:~ q(V).[1@1,V,r2]
\end{verbatim}
}

As the optimality of this answer set is $5$, we know that this hypothesis
cannot have any violating reasons, as otherwise, there would be an answer set
with optimality $4$ corresponding to the hypothesis (which would have been
returned as optimal). Hence, the hypothesis must be an optimal inductive
solution of the task.

\begin{figure*}[ht!]
  \begin{multicols}{2}
{\small
\begin{verbatim}
% meta(VI)
in_vs(p(1),v1).
in_vs(p(2),v1).
in_vs(r(1),v1).
in_vs(r(2),v1).
in_vs(a,v1).
vs(v1).
:- not nas(v1).

% meta(VP)
in_vs(p(2),v2).
in_vs(q(1),v2).
in_vs(r(1),v2).
in_vs(r(2),v2).
in_vs(a,v2).
vs(v2).

in_vs(q(1),v3).
in_vs(q(2),v3).
in_vs(r(1),v3).
in_vs(r(2),v3).
in_vs(a,v3).
vs(v3).

dom_lv(v2,v3,L) :- lv(L),
   #sum{w(W,L,A,v2)=W,
        w(W,L,A,v3)=-W} < 0.
wrong_dom_lv(v2,v3,L) :- lv(L),
   #sum{w(W,L,A,v3)=W,
        w(W,L,A,v2)=-W} < 0.
wrong_bef(v2,v3,L) :- lv(L), L < L2,
   wrong_dom_lv(1,2,L2).
dom(v2,v3) :- dom_lv(v2,v3,L),
   not wrong_bef(v2,v3,L).

:- not nas(v2), not nas(v3),
   not dom(v2,v3).

% meta(Aux)
mmr(p(V),X) :- mmr(r(V),X),
  not in_vs(q(V),X), vs(X).
mmr(q(V),X) :- mmr(r(V),X),
  not in_vs(p(V),X), vs(X).
mmr(r(1),X) :- vs(X).
mmr(r(2),X) :- vs(X).
mmr(a,X) :- not in_vs(b,X), vs(X).
mmr(b,X) :- not in_vs(a,X), vs(X).

mmr(q(1),X) :- vs(X), in_h(r1).
w(1,1,ts(V),X) :- vs(X),
  in_vs(q(V),X), in_h(r2).
w(1,1,args(b,r3),X) :- vs(X),
  in_vs(b,X), in_h(r3).

nas(X) :- in_vs(A,X), not mmr(A,X).
nas(X) :- not in_vs(A,X), mmr(A,X).
\end{verbatim}
}
\end{multicols}
\caption{An example of $VR_{meta}(T)$.\label{fig:vmeta}}
\end{figure*}
\end{example}
\end{appendix}

\end{document}